\documentclass[letterpaper, 10 pt, conference]{ieeeconf} 
\usepackage{amsmath}
\usepackage{url}
\usepackage{graphics}
\usepackage{graphicx}
\usepackage{caption}
\usepackage{subcaption}
\usepackage{algorithmic}
\usepackage{algorithm}
\usepackage{setspace}
\usepackage{epsfig}
\usepackage{amsfonts}
\usepackage{todonotes}

\usepackage{color}
\definecolor{CommentRed}{rgb}{0.7,0,0}
\definecolor{CommentBlue}{rgb}{0,0,0.7}
\definecolor{CommentGreen}{rgb}{0,0.7,0}

\definecolor{myblue}{RGB}{0,51,102}

\IEEEoverridecommandlockouts                              
\title{\LARGE \bf Watch This: Scalable Cost-Function Learning \\ for Path Planning in Urban Environments }

\author{Markus Wulfmeier$^{1}$, Dominic Zeng Wang$^{1}$ and Ingmar Posner$^{1}$
\thanks{$^{1}$The authors are with the Mobile Robotics Group, Department of Engineering Science, 
        University of Oxford, United Kingdom; \newline 
        {\tt\small markus, dominic, ingmar@robots.ox.ac.uk}}%
}

\begin{document}

\maketitle
\thispagestyle{empty}
\pagestyle{empty}

\maketitle

\begin{abstract}
In this work, we present an approach to learn cost maps for driving in complex urban environments from a very large number of demonstrations of driving behaviour by human experts. The learned cost maps are constructed directly from raw sensor measurements, bypassing the effort of manually designing cost maps as well as features. When deploying the learned cost maps, the trajectories generated not only replicate human-like driving behaviour but are also demonstrably robust against systematic errors in putative robot configuration. To achieve this we deploy a Maximum Entropy based, non-linear IRL framework which uses Fully Convolutional Neural Networks (FCNs) to represent the cost model underlying expert driving behaviour. Using a deep, parametric approach enables us to scale efficiently to large datasets and complex behaviours by being run-time independent of dataset extent during deployment. We demonstrate the scalability and the performance of the proposed approach on an ambitious dataset collected over the course of one year including more than 25k demonstration trajectories extracted from over 120km of driving around pedestrianised areas in the city of Milton Keynes, UK. We evaluate the resulting cost representations by showing the advantages over a carefully manually designed cost map and, in addition, demonstrate its robustness to systematic errors by learning precise cost-maps even in the presence of system calibration perturbations.
\end{abstract}

\section{Introduction}
\label{sec:introduction}

The majority of state-of-the-art motion planning systems for autonomous driving applications rely on manually designed cost-functions~\cite{choset2005principles}, with recent successful examples given by the competing teams in the DARPA Grand~\cite{Thrun2006} and Urban Challenges~\cite{urmson2008autonomous,montemerlo2008junior}.  
When designing a cost-function, obstacles typically are inflated as a function of the vehicle size. The weighting of costs from different sensing modalities relies on extremely detailed domain knowledge. In addition, designing good features to extract from raw input data for computing the cost maps is often a non-trivial task relying heavily on a well-understood hardware setup. 

The requirement for high-capacity models for cost-functions arises when one considers the application of planning frameworks in urban environments that are of significant complexity. For example, consider a light-weight electric vehicle designed to transport people in a city between popular locations such as the train station and shopping centres. Because of the proximity to people and the low speed of the vehicle, it mainly operates on pedestrian walkways and cycle paths. This scenario introduces new challenges to the planning framework. In addition to coping with conventional obstacles in usual urban driving scenarios, such as trees, cars and pedestrians, the planner is faced with additional, unconventional obstacles, such as bollards, narrow underpasses and steep ramps that are easily navigated by pedestrians, but challenging for a robot.

Manually designing cost-functions that robustly handle these added complexities is a challenging and time-consuming task. This motivates our approach to learn end-to-end cost-mappings sensory perception based on large amounts of expert demonstrations. Furthermore, this approach provides significant robustness towards systematic inaccuracies that can be found e.g. in system calibration, consequently rendering it more independent of exact knowledge of vehicle configuration.

\begin{figure}[t]
	\centering
	\includegraphics[width=0.48\textwidth]{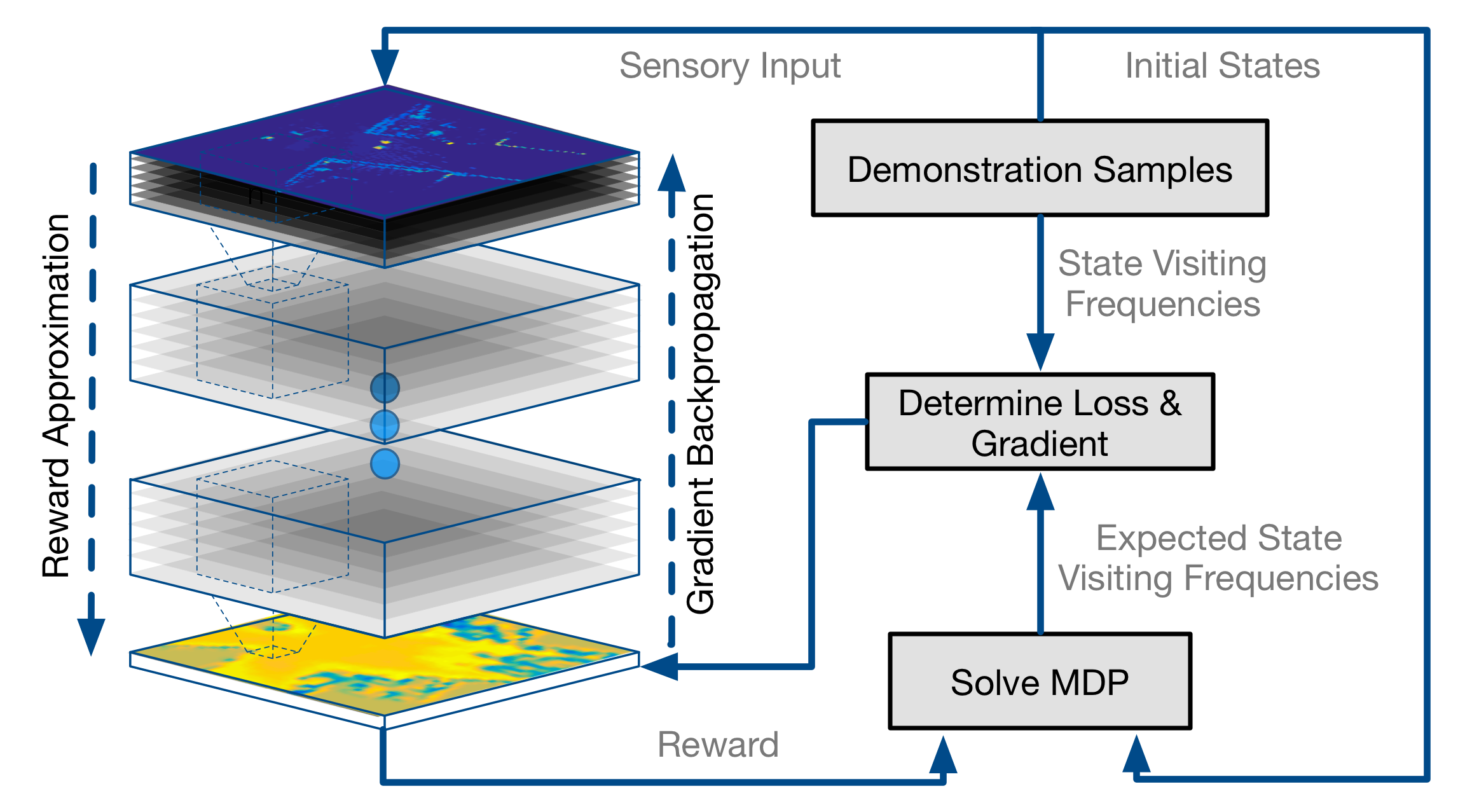}
	\caption{Schema for training neural networks in the Maximum Entropy paradigm for IRL.}
	\label{fig:reward_approx}
\end{figure}

In this work, we formulate cost-function learning from expert demonstrations as an Inverse Reinforcement Learning (IRL) problem~\cite{abbeel2004apprenticeship}. Recently, Wulfmeier et al.~\cite{wulfmeier2015dirl} proposed a framework that introduced training of deep neural networks into the paradigm of Maximum Entropy IRL~\cite{ziebart2008maximum}. The method is based on iterative refinement of the cost-model interwoven with the solution of the planning problem formulated as a Markov Decision Process (see Figure~\ref{fig:reward_approx} for an illustration of the process). Representing the cost-function with a deep architecture is attractive to us because it opens up the possibility of learning high-capacity, highly non-linear models that are necessary for describing complex, real-world urban environments. However, while the original proof-of-concept work targeted the feasibility as well as performance on toy scenarios, the proposed network architecture as it stands does not have enough capacity for our application at hand. In this work, we scale up the framework proposed in~\cite{wulfmeier2015dirl} to cope with the full complexity of real-world urban driving. To capture this complexity, we need to deploy high-capacity models, and in order to train them, we require an extensive amount of demonstration data from real, human drivers. There are several advantages of this approach. First, by learning from human drivers, we learn cost-functions that generate trajectories that mimic human driving, closer to what a passenger will expect. Second, as the proper behaviours at certain types of environments are shown by human experts, we may learn to automatically correct for any systematic biases (such as a misaligned calibration), or traverse an otherwise untraversable path (e.g.\ through a tight pair of bollards, where a conventional planner will have no choice but be conservative). Lastly, deploying a deep neural network enables us to learn good feature representations directly from raw sensor input.

The key contributions of this paper are three-fold:
\begin{enumerate}
    \item New architectures for the Maximum Entropy Deep IRL framework~\cite{wulfmeier2015dirl} that enable it to learn high-capacity, non-linear models that are necessary for handling the complex environments encountered in real-world urban driving scenarios.
    \item A demonstration of the scalability of the proposed approach on a dataset collected over the course of one year capturing the natural driving behaviours of 13 different human drivers, resulting in over 25,000 training samples.
    \item A demonstration of the robustness of the proposed approach to systematic biases in robot configuration, such as sensor miscalibration.
\end{enumerate}

To show the efficacy of the proposed approach, we also compare it against a carefully handcrafted cost-function for path planning that is currently deployed on our research platforms. As the evaluation on our year-long dataset in Section~\ref{sec:experiments} shows, the proposed approach outperforms significantly the hand-designed cost-function. In addition we show that, even under situations of systematic biases such as sensor miscalibration, the proposed approach remains functional, while the hand-designed cost-function fails catastrophically.

\section{Related Works}

Recent works in applying learning from demonstration (LfD) to robotic tasks have led to great advances in different approaches including direct policy learning and IRL. While we mention a select number of relevant works, the interested reader is directed to \cite{argall2009survey} for a detailed summary of work in this area. A main distinction in LfD is found in the type of model that is derived. Policy imitation, also known as behavioural cloning, targets the direct learning of a policy mapping from perceived environment or preprocessed features to the agent's actions. Inverse Reinforcement Learning in contrast focuses on inferring the agent's underlying reward structure.

A principal challenge for policy learning from demonstration lies in generalisation since policies are learned along trajectories and deviations from those introduce large errors. DAgger \cite{ross2010reduction} addresses this problem by querying an expert to improve the policy for states not encountered in the original trajectories but leads to increased and repeated human effort.

While direct policy imitation enables the design of reactive controllers, long-term decision making systems aiming to find a safe, collision-free path in cluttered terrains need to plan further ahead and test plans against constraints. 
Furthermore, a reward model is generally seen to be more succinct than the policy and conceived to be preferable as generalisation becomes important \cite{abbeel2004apprenticeship,abbeel_apprenticeship_2008}.

Existing work in Inverse Reinforcement Learning shows the ease of directly deploying the learned cost maps with existing state-of-the-art planning systems  \cite{ratliff_learning_2009,ratliff_imitation_2007,jain2015planit}, while current policy imitation based approaches \cite{pomerleau1990neural,sermanet_multi-range,chen2015deepdriving} are of limited use in long-term decision making tasks.

Due to its strengths for planning tasks and the capability for integration into existing systems, recent work on IRL lead to significant success for long-term planning tasks. Results include improving driving and robotic navigation with focus on interaction of mobile autonomous platforms and humans \cite{kuderer2015learning,kretzschmar2016socially}. Jain et al.~\cite{jain2015planit} addressed learning from direct user interaction focusing on scaling to large user bases of non-experts. 
In contrast to these works, we develop an approach that scales with both the size of the training set and additional complexity in the input data by applying Fully Convolutional Neural Networks (FCNs)~\cite{long2015fully} with the principal goal of a direct perception-to-cost mapping and robustness towards systematic biases in the configuration of the robot.

While recent work on Expectation-Maximisation for learning multiple reward functions \cite{NIPS2015_5882} presents a promising approach to handle multiple demonstrators, our principal goal is to learn a model of the common patterns between all drivers and approximating a general human reward function for driving.

Maximum Entropy IRL \cite{ziebart2008maximum} represents a state-of-the-art approach to learning reward mappings by framing the demonstration trajectories as drawn from a distribution with the probability for a trajectory only depending on the expected future rewards. While the method addresses expert suboptimality in an efficient way it only provides a linear reward model that can easily be overburdened with approximating complex reward functions. 
In parallel to this work, Finn et al developed a sampling based approach for IRL to train non-convolutional neural networks in a MaxEnt framework for tasks of robotic manipulation and navigation \cite{FinnLA16}. 

Non-parametric methods such as Gaussian Processes (GPs) have been employed to overcome these limitations~\cite{levine2011nonlinear} and while this in principle extends the IRL paradigm to the flexibility of nonlinear reward approximation, the use of a nonparametric model makes it prone to requiring a large number of demonstration samples in order to approximate highly varying reward functions \cite{bengio2007scaling}. 
The situation targeted in this work leads to complex reward mappings based directly on sensory data without manual feature design and therefore high demands regarding to the dataset size which quickly renders any nonparametric approach computationally impracticable.

Our work addresses complex, non-linear reward functions by applying FCNs to learn necessary feature representations and reward functions for IRL \cite{wulfmeier2015dirl} and extends existing work towards the application in a large scale robotics scenario with tens of thousands of demonstration trajectories.

\section{Methodology}

We address the task of learning a cost-function for the motion planning system by applying deep learning to the Maximum Entropy paradigm for IRL. The resulting training algorithm focuses on the ideas of dynamic programming and backpropagation. Compared to early work \cite{wulfmeier2015dirl}, we extend the neural network architecture to handle features at multiple scales using parallel information streams based on a directed acyclic graph.

\subsection{Inverse Reinforcement Learning}
\label{sec:irl}
The principal goal of IRL is to infer the preferences underlying specific behaviours. The task is often solved based on framing the task in the Markov Decision Process (MDP) framework, which can be defined as $\mathcal{M} = \{ \mathcal{S}, \mathcal{A}, \mathcal{T}, \gamma, r \}$, where $\mathcal{S}$ denotes the state space, $\mathcal{A}$ denotes the set of possible actions, $\mathcal{T}$ denotes the transition model, $\gamma$ is the discount factor - a value in $(0,1]$ reducing the influence of future rewards - and finally $r$ is the reward structure.
The terms ``cost" and ``reward" are used interchangeably in the following sections since one can be converted into the other by negation.

IRL specifically addresses the case where, instead of the reward structure, a set of expert demonstrations $\mathcal{D} = \{ \varsigma_1, \varsigma_2, ..., \varsigma_N \}$ are supplied. Each demonstration consists of a sequence of state-action pairs such that $\varsigma_i = \{(s_1,a_1),(s_2,a_2),...,(s_K,a_K)\}$. The principal goal is to infer the underlying reward $r$ of the expert demonstrations, which can be used either to predict behaviour or even to replicate it. 

The general formulation of IRL results in two main complications. The expert has to be modelled in a way that suboptimality with respect to the reward can be handled. Furthermore, additional constraints have to be introduced to dissolve the inherent ambiguity of reward structures for given demonstrator behaviour \cite{abbeel2004apprenticeship}. For this work, the robust handling of suboptimality is of absolute necessity since the demonstration trajectories are extracted from large driving datasets without specified driving behaviour and multiple drivers as further explained in section \ref{sec:experiments}.

\subsection{Maximum Entropy Deep Inverse Reinforcement Learning}
\label{sec:training}

The Maximum Entropy paradigm (MaxEnt) for IRL addresses the suboptimality and reward ambiguity problems mentioned in section \ref{sec:irl} by modelling expert behaviour as a distribution over expert trajectories and constraining this distribution to the one of highest entropy~\cite{ziebart2008maximum}. 
Furthermore, as shown in~\cite{wulfmeier2015dirl}, it leads to a fully differentiable objective function, thus enabling backpropagation of loss gradients and lends itself naturally for training neural network architectures. We formulate the training procedure in what follows as a straightforward stochastic gradient descent optimisation.

The MaxEnt formulation defines the agent's policy $\pi_D(a|s)$ such that 
the resulting behaviour is directed towards the maximisation of rewards given the current model. This leads to the probability for user preference of any given trajectory between specified start and goal states being proportional to the  exponential of the reward along the path:
\begin{equation}
    P(\varsigma|r) = \prod_{i=1}^K \pi_D(a_i|s_i) \propto \exp \{ \sum_{{i=1}}^K  ~ r_{s_i,a_i} \}.
\label{eq:maxent_prob}
\end{equation}

The complete objective for Maximum Entropy Deep Inverse Reinforcement Learning \cite{wulfmeier2015dirl} is based on a data term, maximising the likelihood of demonstration data given the parametrised reward function, as well as a model term for regularisation purposes:
\begin{equation}
\mathcal{L(\theta)} = \log P(\mathcal{D},\theta|r(\theta)) = \underbrace{\log P(\mathcal{D}|r(\theta))}_{\mathcal{L_D}} + \underbrace{\log P(\theta)}_{\mathcal{L_\theta}}.
\label{eq:maxent_objective}
\end{equation}
Examples of useful regularisers include the $l_1$ and $l_2$ norms or a combination of both (Elastic Net~\cite{Zou05regularizationand}).

The data-based gradient given by the Maximum Entropy approach can be separated into two gradients by the application of the chain rule -- the gradient of the objective with respect to the reward and the gradient of the reward with respect to the network parameters \cite{wulfmeier2015dirl}:
\begin{eqnarray}\label{eq:maxent_loss}
\frac{\partial \mathcal{L_D}}{\partial \theta} &=& \frac{ \partial \mathcal{L_D} }{ \partial r } \frac{ \partial r }{ \partial \theta }\\
&=& \underbrace{(\mu_{\mathcal{D}} - \mathbb{E} [\mu])}_{\mbox{\small MaxEnt}} \underbrace{\frac{\partial}{\partial \theta} r(\theta)}_{\mbox{\small Backpropagation}} .
\end{eqnarray}
By extending the original linear formulation to neural networks, we benefit from the efficiency of gradient backpropagation to associate the difference in state visitation frequencies $\mu$ with the most relevant parameters.

Algorithm \ref{alg:medirl} shows the procedure for iterative refinement of the reward model based on gradients computed with Equation \ref{eq:maxent_loss}, while Figure \ref{fig:reward_approx} intuitively visualises the training procedure. The expert's state visitation frequencies $\mu_D$ represent the number of visits to a specific state extracted from the training data and represents the sum over actions for the state-action visitation frequencies $\mu_D^a$.
The learner's expected state visitation frequencies $\mathbb{E} [\mu]$ are being calculated following Algorithm \ref{alg:medirl}. The input to the neural network (the sensor measurement) is represented by $f$.
A more detailed illustration of the algorithm including the dynamic programming formulation for value iteration and policy propagation in lines 4 \& 5 can be found in \cite{wulfmeier2015dirl}.

\begin{algorithm}[h]
	\caption{Maximum Entropy Deep IRL}  
		\begin{algorithmic}[1]\label{alg:medirl} 
		    \begin{spacing}{1.5}
			\REQUIRE $\mu_D^a, f, S, A, T, \gamma$
			\ENSURE $\mathrm{optimal ~ weights}~ \theta^*$ 
			\vspace{1mm}
			\STATE{$\theta^1 = \mathrm{initialise\_weights()}$} \\
			\vspace{3mm}
			\textcolor{myblue}{\textbf{Iterative model refinement}}\\
			\FOR{n = 1 : N} 
			\STATE{$r^n = \mathrm{nn\_forward}(f,\theta^n)$ } \\
			\vspace{3mm}
		    \textcolor{myblue}{{\textbf{Solution of MDP with current reward}}}\\
			\STATE{$\pi^n = \mathrm{approx\_value\_iteration}(r^n,S,A,T,\gamma)~~~~~~~~~$} \\ 
			\STATE{$\mathbb{E} [\mu^n] = \mathrm{propagate\_policy}(\pi^n,S,A,T)~~~~$}  \\ 
			\vspace{3mm}
			\textcolor{myblue}{{\textbf{Maximum Entropy loss and gradients}}}\\
			\STATE{$\mathcal{L}_D^n = \log(\pi^n) \times \mu_D^a$ }\\
			\STATE{$\frac{\partial \mathcal{L}_D^n}{\partial r^n}  = \mu_D - \mathbb{E} [\mu^n]$ }\\
			\vspace{3mm}
			\textcolor{myblue}{{\textbf{Compute network gradients}}}\\
			\STATE{$\frac{\partial \mathcal{L}_D^n}{\partial \theta_D^n} = \mathrm{nn\_backprop}(f, \theta^n, \frac{\partial \mathcal{L}_D^n}{\partial r^n}) ~~$}  \\ 
			\STATE{$\theta^{n+1} = \mathrm{update\_weights}(\theta^n, \frac{\partial \mathcal{L}_D^n}{\partial \theta_D^n} ) $}
			\ENDFOR
			\end{spacing}
		\end{algorithmic}
\end{algorithm}

\subsection{Multi-Scale Architectures}
\label{sec:deeparch}

\begin{figure}
	\centering
	\includegraphics[width=0.48\textwidth]{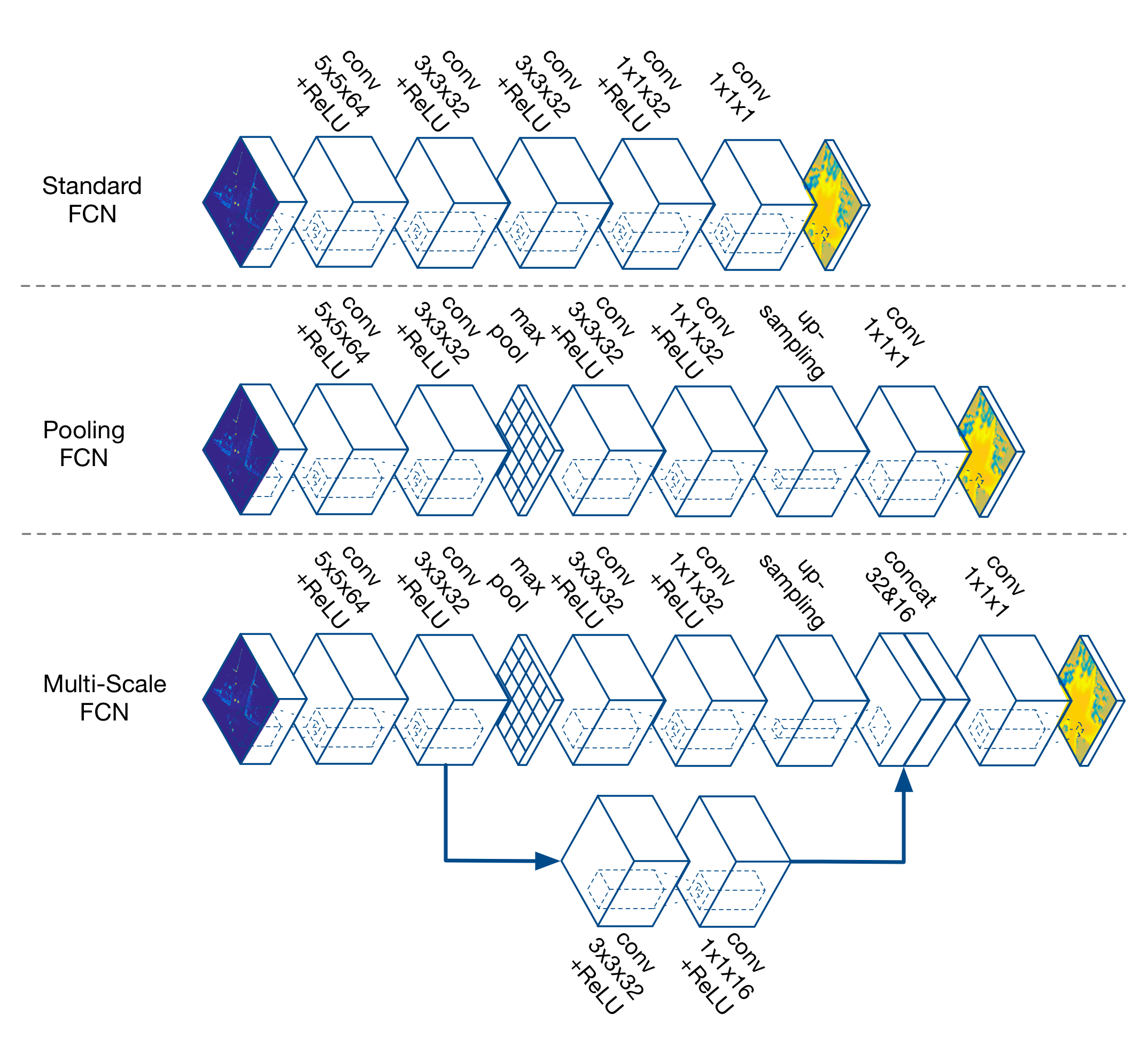}
	\caption{Illustration of the three proposed network architectures.}
	\label{fig:arch}
\end{figure}

In addition to the basic Fully Convolutional Network (FCN) presented in \cite{wulfmeier2015dirl} a main contribution of this work is the design of two extended architectures based on advances in other applications for neural networks to address shortcomings of the original network.

The pooling FCN illustrated in Figure \ref{fig:arch} simply introduces a notion of translation invariance and reduces the size of the representation in the following layers. The extension is needed to address the influence of features on the following layers with limited spatial invariance. Especially in the application in section \ref{sec:experiments}, the areas further away from the car cannot be densely sensed via LIDAR and the resulting patterns in all feature channels can be hard to model in all possible variations. While a larger number of low level filters might be able to address the issue, it will increase the chance of overfitting. Max-Pooling layers introduce a limited invariance with respect to translations which reduces the modelling efforts and increases accuracy without increasing the chance of overfitting.

While the pooling based architecture reduces the spatial size of the representation and leads to increased spatial invariance, this also leads to the loss of location information for low level features. Inspired by work on the multi-scale Deep Jet architecture \cite{long2015fully} that integrates features of different scales, we introduce the Multi-Scale (MS) FCN architecture shown in Figure \ref{fig:arch}. The difference between our proposed MS FCN architecture and the Deep Jet architecture is that, instead of summing the contributions from the two branches of the network, we concatenate them following~\cite{Szegedy_2015_CVPR}. This modification to the Deep Jet architecture is important because in our case the semantic meanings of the feature channels from the two branches differ. Hence to prevent mixing of the two channels and to retain all information we concatenate them. This is in contrast to the Deep Jet architecture~\cite{long2015fully}, where feature channels from all branches have the same semantic meaning, hence it is meaningful to add them.

The MS FCN architecture addresses the influence of features of different scales on the reward as well as the possibility for spatial invariance with high level features while retaining the precise location information offered by the low level features.
Furthermore, all architectures are designed such that the size of the receptive field for each location in the final cost-map is ensured to be large enough to encapsulate the area occupied by the vehicle plus additional surrounding terrain.

In the next section, we evaluate all three architectures qualitatively and quantitatively, and compare them with a carefully manually designed cost-function that is currently deployed on our research platforms.

\section{Experiments}
\label{sec:experiments}

\subsection{Large-Scale Data Collection}

We demonstrate the applicability of the proposed approach to complex, real-world, urban driving environments on a large dataset that was collected over the course of one year involving 13 different drivers driving manually every two weeks on the pedestrian walkways and cycle lanes in the city of Milton Keynes, UK.

\begin{figure}[h]
	\centering
	\includegraphics[width = 0.45\textwidth]{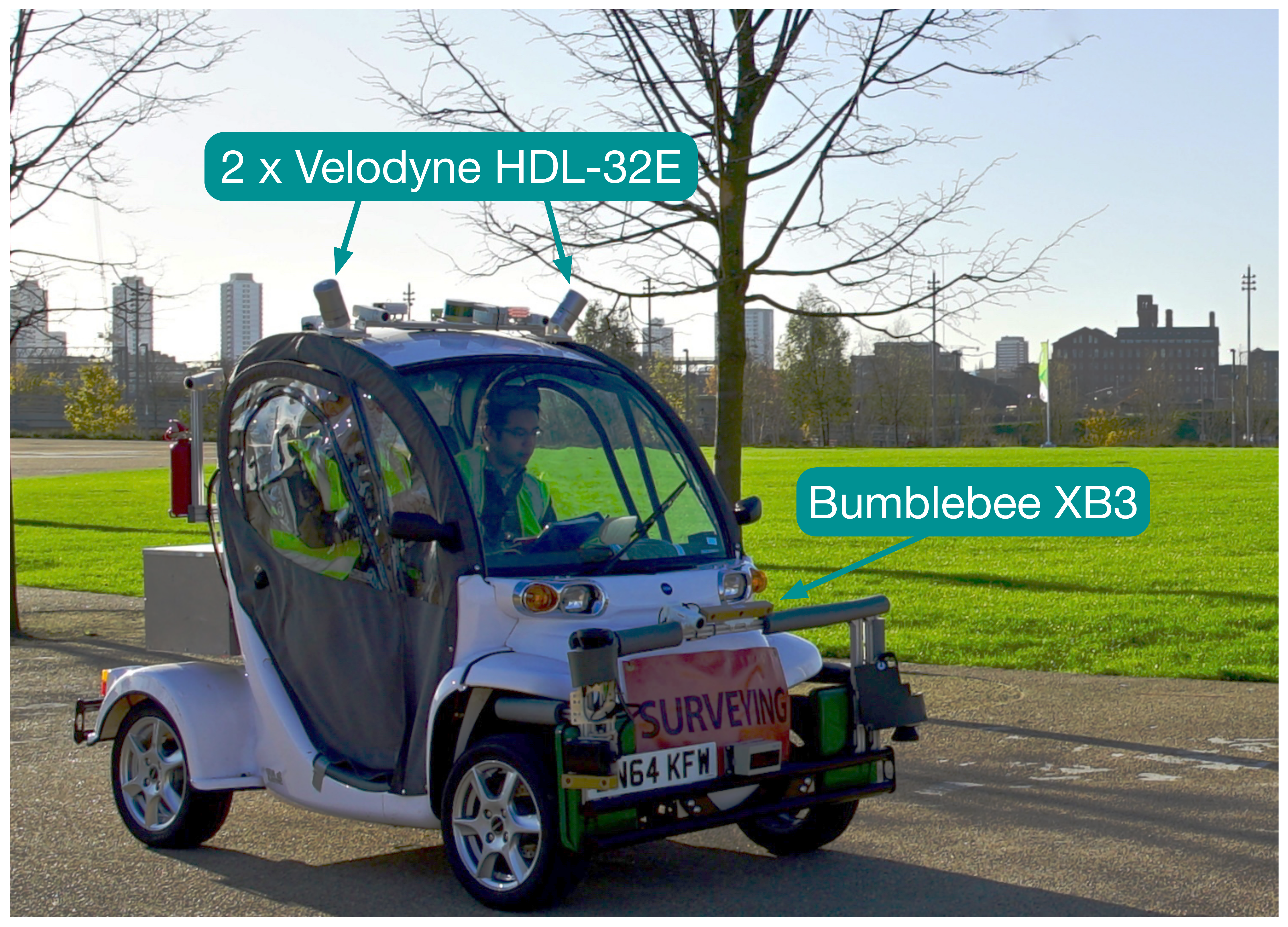}
	\caption{Our research mobile platform -- a modified GEM golf cart.}
	\label{fig:milton}
\end{figure}

Our data collection platform is a modified GEM (Global Electric Motorcars) golf cart (cf.\ Figure~\ref{fig:milton}),
equipped with a variety of sensors including 2D as well as 3D LIDARs and stereo- and mono-cameras. The relevant sensors for this work include two Velodyne HDL-32E scanners and a Bumblebee XB3 stereo camera.

The dataset collected contains a variety of challenging obstacles such as trees, bollards, bike racks, curbs, stairs and pedestrians. An additional challenge of the dataset lies in the variation in driving behaviours and styles due to the multiple drivers. 

We extracted in total over 25,000 trajectories, each about 15m long, from the more than 120km of driving contained in the dataset. The extracted trajectories are then randomly divided into a training set containing 95\% of the data and a test set based on the remaining trajectories.

\subsection{Input Representations}
The input data to our network is based on pointclouds measured by the two 3D Velodyne scanners with overlapping field-of-view. Given the calibration of the LIDARs to the robot frame, the resulting 3D pointclouds are mapped into a 2D-grid based static map on the ground plane to enable the application of FCNs to the environment representation. The extracted statistics include mean height, height variance and a binary indicator if the cell is visible in any scan. The grid has a size of 25m x 25m and a resolution of 0.25m per cell.

\subsection{Demonstration Trajectories}
\label{sec:vo}

\begin{figure}
	\centering
    \begin{subfigure}[b]{0.15\textwidth}
		\includegraphics[width = \textwidth]{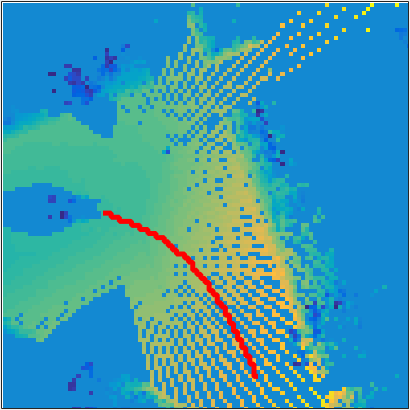}
	\end{subfigure}
	\begin{subfigure}[b]{0.15\textwidth}
		\includegraphics[width = \textwidth]{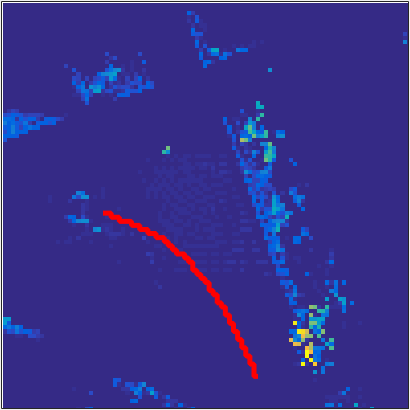}
	\end{subfigure}
	\begin{subfigure}[b]{0.15\textwidth}
		\includegraphics[width = \textwidth]{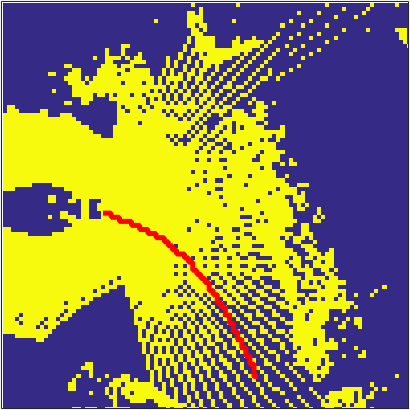}
	\end{subfigure}
	\caption{Visualisation of demonstration trajectories (in red) on all fields of a static map. From left to right: mean height, height variance, cell visibility. }
	\label{fig:training_samples}
\end{figure}

Figure~\ref{fig:training_samples} shows an example of gathered trajectories based on demonstration driving data. As a first step a serial chain of transforms is extracted from the motion of the robot estimated from Visual Odometry~\cite{churchill2012continually} on stereo images from the Bumblebee XB3 camera. The extracted transform chain is subsequently mapped into the static map frame and discretised to fit into the grid representation. Repeated states are removed from the sequences before conversion into state-action trajectories through the MDP to prevent overemphasis of keep-position actions. Finally, the action space is simplified to a discrete set of motions around the current position. 
The simplification is necessary to keep the problem tractable and use the given environment representation based on a 2D grid necessary for convolutions.

\subsection{Implementation Details}
\label{sec:details}

Following each non-linearity we insert a layer of Batch Normalisation (BN)~\cite{ioffe2015batch} to reduce internal co-variate shift within batches and speed up training, while adding to model regularisation (we use Elastic Net~\cite{Zou05regularizationand} for all our experiments) and increasing generalisation performance.
While the introduction of Dropout \cite{hinton2012dropout} in low rates into the network can generally lead to higher robustness to noise and overfitting even in small networks, early experiments showed that the use of BN lead to significantly faster training and similar performance in the converged models. A combination of both would require to use Dropout in the testing phase as the BN parameters are adapted to handle the training Dropout rates, but this approach leads to significantly slower run-times as a larger number of Monte-Carlo samples would have to be calculated for a reliable estimate of the mean reward.

To increase the number of training samples, the set includes partially overlapping trajectories with static maps from different positions. While the trajectory samples can be seen as not fulfilling the i.i.d.\ criterion, the advantage of additional input representations from different view points -- the static maps -- leads to significant performance gain after randomisation of the data sequences. A principal reason for this fact is that feature representations based on LIDAR scans differ significantly according to the distance away from the sensor as laser points become sparser as the distance from the sensor increases. Hence different viewpoints of the same environment from multiple static maps help in this situation.

\subsection{Evaluation}
\label{sec:predict}

To evaluate the accuracy of approximating human driving behaviour, we use two metrics: the negative log-likelihood of the demonstration data (NLL) as well as the Modified Hausdorff Distance (MHD) \cite{kitani_activity_2012}. Both metrics evaluate model performance for predicting demonstrator behaviour on the test set with the latter being used to determine a metric of distance between the original demonstration trajectories and new samples generated based on the cost-function and resulting policy of the trained system.
The presented metrics result from averaging over the metrics for 1000 test trajectories and in case of the MHD 10 samples per single test.

\begin{figure}[h]
	\centering
		\includegraphics[width = 0.2\textwidth]{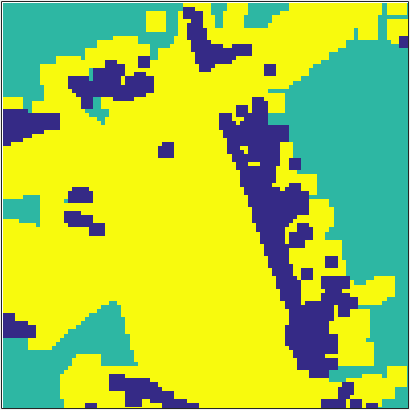}
	\caption{Handcrafted cost-function (our baseline for comparison) for the same scenario shown in Figure \ref{fig:training_samples}.}
	\label{fig:manual_cost-map}
\end{figure}

As part of the following sections, we evaluate the different models against the handcrafted cost-function currently applied in the planning module. The main idea behind the design of this cost-function, as illustrated in Figure \ref{fig:manual_cost-map}, is that obstacles generally represent areas of large height variance in each cell and areas without LIDAR information are labelled as unknown, which in the classification task is treated as not traversable. Furthermore, obstacles are extended by the radius of the smallest circle encapsulating the GEM platform, an approach commonly known as Minkowski sum \cite{lavalle2006planning}.

\begin{table}[h!]
\begin{center}
\begin{tabular}{ |p{0.1\textwidth}||p{0.06\textwidth}|p{0.06\textwidth}|p{0.06\textwidth}|p{0.06\textwidth}|  }
     \hline
     Prediction metrics &  Standard FCN & Pooling FCN &  MS \newline FCN & Manual CF\\
     \hline
     \hline
     NLL & \hfill 69.35 & \hfill 69.73 & \hfill 65.39 & \hfill 78.13\\ 
     \hline
     MHD   & \hfill 0.221 & \hfill 0.230 & \hfill 0.200 & \hfill 0.284\\             
     \hline
\end{tabular}
\caption{Prediction performance on the test set given by Negative Log-Likelihood (NLL) and Modified Hausdorff Distance (MHD).}
\label{tab:eval}
\end{center}
\end{table}

\begin{figure}[h]
	\centering
    \begin{subfigure}[b]{0.15\textwidth}
		\includegraphics[width = \textwidth]{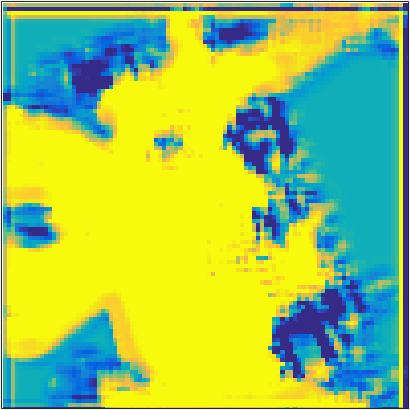}
		\caption{}
	\end{subfigure}
	\begin{subfigure}[b]{0.15\textwidth}
		\includegraphics[width = \textwidth]{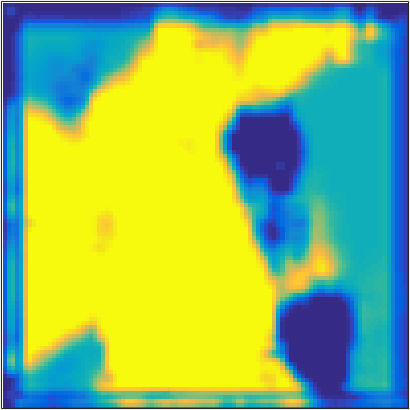}
		\caption{}
	\end{subfigure}
	\begin{subfigure}[b]{0.15\textwidth}
		\includegraphics[width = \textwidth]{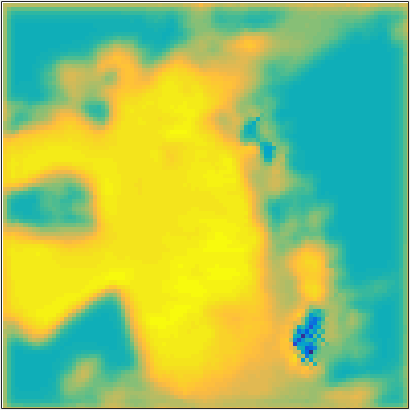}
		\caption{}
	\end{subfigure}

	\caption{Visualisation of the learned cost-maps based on (a) Standard FCN, (b) Pooling FCN, and (c) MS FCN.}
	\label{fig:learned_cost-maps}
\end{figure}

While the standard FCN already predicts test trajectories significantly better than the manual cost-function (see Table \ref{tab:eval}) and therefore describes human driving behaviour more accurately, the qualitative analysis of the demonstration maps in Figure~\ref{fig:learned_cost-maps} shows that for areas where LIDAR scans get sparser the network has problems reasoning about the true reward.
The use of the pooling architecture without parallel information branches performs better in terms of a smooth cost-function when interpreting the same areas. Due to the Max-Pooling layer the approach is able to infer traversable terrain. However, it also leads to a loss of information in the pooling step as discussed in section \ref{sec:deeparch}.
Best performance is achieved when spatial invariance in one branch is combined with the preservation features in a parallel chain of the DAG-like MS-FCN architecture.

We furthermore perform a quantitative analysis of the resulting cost model by evaluating the binary classification of trajectories as collision/free space with the results presented in Table \ref{tab:classif}.
Since we try to learn the unspecified cost-function that underlies the mutual behaviour of multiple drivers, there is no ground-truth in costs. However, the driven trajectories themselves present knowledge about the actual environment and every driven trajectory is clearly free of collisions. 

The classification of trajectories based on learned cost-functions needs a threshold to decide for the minimal cost at which a collision occurs and the trajectory can be classified as untraversable. To enable the determination of such a threshold we introduce artificial trajectories representing collisions into the evaluation process such that we can determine Type I (false positives) and Type II errors (false negatives) for the classification. The artificial trajectories were chosen based on direct information from LIDAR as well as camera data that ensures collision.
The final decision for a threshold was made to reduce FPR to 0\%. It can be preferable to increase the threshold slightly such that the FNR drops significantly and the cost-function enables planning to find more traversable paths. This leads however to an increase in FPR and should be constrained by setting a maximum limit.

\begin{table}[h!]
\begin{center}
\begin{tabular}{ |p{0.1\textwidth}||p{0.06\textwidth}|p{0.06\textwidth}|p{0.06\textwidth}|p{0.06\textwidth}|  }
     \hline
     Error metrics &  Standard FCN & Pooling FCN &  MS \newline FCN & Manual CF\\
     \hline
     \hline
     FNR   & \hfill 0.471 & \hfill 1.000 & \hfill 0.206 & \hfill 0.441\\
     \hline  
     FPR & \hfill 0.000 & \hfill 0.000 & \hfill 0.000 & \hfill 0.000\\
     \hline
\end{tabular}
\caption{Trajectory evaluation performance.}
\label{tab:classif}
\end{center}
\end{table}

With the manual cost-function design being strictly conservative with respect to the traversability of terrain it results in 0 \% false positives rate (FPR) while showing a significant number of false negatives. The adaption of cost-function thresholds for 0\% FPR turns out to be complicated. While it is often easy to achieve rates of under 5\% the final adjustment significantly increases the false negatives rate. In case of the pooling architecture this rendered it infeasible to continue planning with 0\% FPR, since a FNR of 100\% represents that none of the paths are classified as traversable. The MS FCN which combines the benefits of both other models (see section \ref{sec:deeparch}) performs best with a reduction in FNR of about 50\% in comparison to the manual cost-map.


\subsection{Robustness Study on Systematic Biases}
\label{sec:robust}

\begin{figure}[h]
	\centering
	\begin{tabular}{c c}
		\includegraphics[width = 0.2\textwidth]{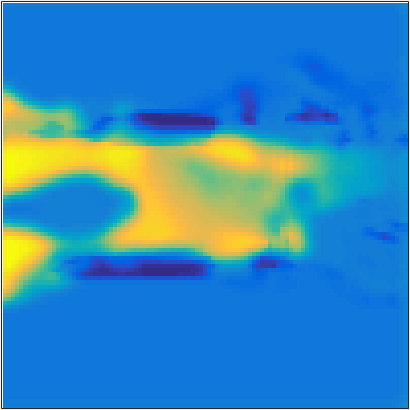}&
        \includegraphics[width = 0.2\textwidth]{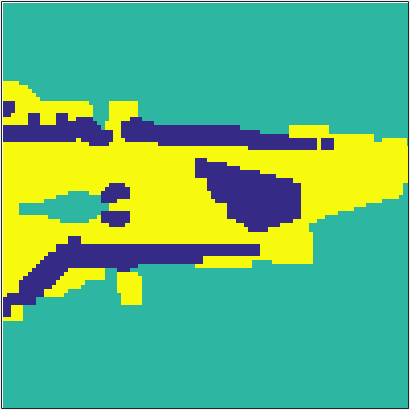}\\
	\end{tabular}
    \caption{Example cost-maps based on miscalibrated data with MS FCN (left), and the handcrafted cost-function (right).}
	\label{fig:robust}
\end{figure}

Learning a cost-function and its most relevant features proves robust towards limited systematic flaws in the configuration of the robot. An inaccurate calibration for example can lead to complete failure for a manually crafted cost-function as presented in Figure \ref{fig:robust}. The artificial obstacles in front of the vehicle can be created due to an imprecise calibration of the pitch angle between the platform and one of the LIDARs of as little as 1$^\circ$. Due to this introduced perturbation in pitch angle of the right Velodyne, the manually defined cost-map creates obstacles in this instance as the height variance of points in a specific cell increases significantly with rising distance from the vehicle.

However, the learned cost-map is able to handle this problem and learn that driving is possible over this terrain. As long as the input representation is rich enough to distinguish between the new features describing artificial obstacles and real walls etc. the system is able to differentiate between features representing obstacles that we have never traversed in any demonstration trajectories and acceptable terrain which was often shown to be traversable in the training data.

Taking the metrics introduced in Section~\ref{sec:predict}, we evaluate the performance of the MS FCN architecture versus the manually defined cost-function, clearly showing the benefits in using a trained system in Tables \ref{tab:robust1} and \ref{tab:robust2}. 

The handcrafted cost-function leads to nearly impassable cost maps in this scenario with an FNR of more than 97\%. The learned model, on the other hand, gives an FNR that, though compares worse than the case with the correct calibration (cf. Table~\ref{tab:classif}), still remains within functional range.

\begin{table}[h!]
\parbox{.45\linewidth}{
\centering
\begin{tabular}{ |p{0.06\textwidth}||p{0.045\textwidth}| p{0.055\textwidth}| }

     \hline
     Error metrics & MS \newline FCN & Manual CF \\
     \hline
     \hline
     FNR   & \hfill 0.525& \hfill 0.971\\
     \hline
     FPR  & \hfill 0.000 & \hfill 0.000\\
     \hline
\end{tabular}
\caption{Trajectory evaluation performance on miscalibrated data.}
\label{tab:robust1}
}
~
\parbox{.45\linewidth}{
\centering
\begin{tabular}{ |p{0.075\textwidth}||p{0.045\textwidth}| p{0.055\textwidth}| }

     \hline
     Prediction metrics &  MS \newline FCN & Manual CF \\
     \hline
     \hline
     NLL   & \hfill 69.35 & \hfill 89.40 \\        
     \hline
     MHD & \hfill 0.267 & \hfill 0.432 \\        
     \hline
\end{tabular}

\caption{Prediction performance on miscalibrated data.}
\label{tab:robust2}
}

\end{table}


\subsection{Discussion}
\label{sec:discuss}

We argue that given the large amount of training data and complex features, only a high-capacity, parametric approach to IRL possesses the capability to approximate the reward mapping while being computationally tractable. Handcrafted features of the required complexity can only be designed with significant expert knowledge of the domain, perception pipeline and expected behaviour. As seen in Section~\ref{sec:experiments}, features describing domain specific obstacles can be easily missed in the preprocessing setup and are inherently hard to design. Spatial features as learned by an FCN on the other hand are inherently optimised for the task. Robustness with respect to unknown but systematic perception inaccuracies as in Section~\ref{sec:robust} is generally beyond reach since the feature construction would depend on knowledge of the type of feature resulting from e.g. miscalibration. Furthermore, while nonparametric approaches such as GPIRL might possess the capacity to approximate complex nonlinear reward functions when given these hypothetically perfect features, the amount of training data severely influences the feasibility of such an approach - potentially rendering it completely intractable. 
While showing great performance on prediction and classification tasks, one consistent aspect of all learned cost-functions represented in Figure~\ref{fig:learned_cost-maps} is the smoothing of the cost-function around obstacles, which is caused since training focuses on features describing traversable terrain. Since most driving in the demonstrations happens around the middle of paths, the features are optimised to represent these areas.
During test time, the agent can encounter a significantly different sensor data distribution. While IRL generalises preferably to direct imitation, this problem can be additionally addressed by transferring a DAgger-like~\cite{ross2010reduction} approach from behavioural cloning to IRL. Furthermore, training with negative examples -- artificial collision trajectories~\cite{kyriacos_inversefailure2015} -- can lead to a generally more expressive cost-function for impassable terrains. 

\begin{figure}[h]
	\centering

    \begin{subfigure}[b]{0.11\textwidth}
		\includegraphics[width = \textwidth]{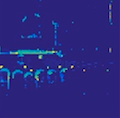}
		\caption{Variance }
	\end{subfigure}
	\begin{subfigure}[b]{0.11\textwidth}
		\includegraphics[width = \textwidth]{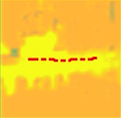}
		\caption{Cost-map}
	\end{subfigure}
	\begin{subfigure}[b]{0.11\textwidth}
		\includegraphics[width = \textwidth]{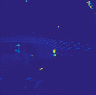}
		\caption{Variance }
	\end{subfigure}
	\begin{subfigure}[b]{0.11\textwidth}
		\includegraphics[width = \textwidth]{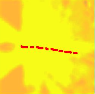}
		\caption{Cost-map}
	\end{subfigure}

    \caption{Example for removal of pedestrians -- represented by bright spots in the height variance map (a\&c) -- from static cost maps (b\&d). }
	\label{fig:ped}
\end{figure}

The current approach focuses on generating a cost map with respect to static environment aspects as represented in Figure \ref{fig:ped}. This is deemed advantageous as the planning module separates into determining spatially feasible paths and subsequent speed-profiling with respect to dynamic obstacles to reduce the dimensionality of the problem.

While this spatio-temporal separation fulfils the requirements of the current motion planning module, future research will address the additional learning of the driver's preference for specific speed profiles in presence of dynamic obstacles. This will require the processing of multiple sensor inputs from different times instead of focusing on from a limited period around the start of each trajectory, since the position of dynamic obstacles will change during the traversal of a given trajectory. A natural extension here would be stacking of these sensor inputs or sequential processing via a recurrent module.

\section{Conclusion and Future Work}

In this work we apply the framework of Maximum Entropy Deep Inverse Reinforcement Learning \cite{wulfmeier2015dirl} to large scale urban navigation scenarios with over 25k demonstration samples collected from 13 different drivers. Based on the extensive dataset of over 120 km of driving, the high capacity of FCNs enables us to deduce the operator's underlying reward mapping directly from sensory input. This is possible even under systematic perturbations of the robot configuration which render handcrafted cost-functions completely impracticable.

We develop a multi-scale network architecture capable of capturing significant features at different scales and are able to exceed the manual cost-function in prediction performance as well as the classification accuracy for test trajectories. 
Due to working directly on the sensory input and learning spatial features, the approach is able to correct for systematic biases which we demonstrate by applying it to a miscalibrated dataset.

Future work will target the investigation of learning from negative demonstration samples (collisions), off-road driving behaviour and extension towards dynamic obstacles as discussed in Section~\ref{sec:discuss}.

\bibliographystyle{unsrt}
\bibliography{main}

\end{document}